\documentclass[11pt]{article}

\usepackage[final]{acl}

\usepackage{times}
\usepackage{latexsym}

\usepackage[T1]{fontenc}

\usepackage[utf8]{inputenc}

\usepackage{microtype}

\usepackage{inconsolata}

\usepackage{graphicx}

\usepackage{lscape}
\usepackage{booktabs}
\usepackage{authblk}

\title{Evaluating LLM-Generated Legal Explanations for Regulatory Compliance in Social Media Influencer Marketing}

\author[1]{Haoyang Gui}
\author[1]{Thales Bertaglia}
\author[1]{Taylor Annabell}
\author[1]{\\Catalina Goanta}
\author[2]{Tjomme Dooper}
\author[3]{Gerasimos Spanakis}

\affil[1]{Utrecht University, The Netherlands}
\affil[2]{Stichting Reclame Code, The Netherlands}
\affil[3]{Maastricht University, The Netherlands}

\begin{document}
\maketitle

\begin{abstract}
The rise of influencer marketing has blurred boundaries between organic content and sponsored content, making the enforcement of legal rules relating to transparency challenging. Effective regulation requires applying legal knowledge with a clear purpose and reason, yet current detection methods of undisclosed sponsored content generally lack legal grounding or operate as opaque ``black boxes.'' Using 1,143 Instagram posts, we compare \textit{gpt-5-nano} and \textit{gemini-2.5-flash-lite} under three prompting strategies with controlled levels of legal knowledge provided. Both models perform strongly in classifying content as sponsored or not (F1 up to 0.93), though performance drops by over 10 points on ambiguous cases. We further develop a taxonomy of reasoning errors, showing frequent citation omissions (28.57\%), unclear references (20.71\%), and hidden ads exhibiting the highest miscue rate (28.57\%). While adding regulatory text to the prompt improves explanation quality, it does not consistently improve detection accuracy. The contribution of this paper is threefold. First, it makes a novel addition to regulatory compliance technology by providing a taxonomy of common errors in LLM-generated legal reasoning to evaluate whether automated moderation is not only accurate but also legally robust, thereby advancing the transparent detection of influencer marketing content. Second, it features an original dataset of LLM explanations annotated by two students who were trained in influencer marketing law. Third, it combines quantitative and qualitative evaluation strategies for LLM explanations and critically reflects on how these findings can support advertising regulatory bodies in automating moderation processes on a solid legal foundation.    
\end{abstract}

\section{Introduction and background}

The rapid rise of social media has made influencer marketing a central strategy for brands seeking to shape followers' purchasing decisions through influencers’ reach and credibility ~\citep{de_veirman_marketing_2017}. While effective at enhancing trust and engagement, this strategy is often opaque, as influencers generally avoid disclosures to maintain authenticity or protect follower engagement. Consequently, sponsored content is frequently hidden or inadequately disclosed~\citep{ershov_effects_2020}, limiting the consumers’ ability to recognise advertising\footnote{This paper uses the terms advertising, sponsored content (posts), and ads interchangeably} and making regulatory oversight difficult.

\begin{figure}
    \centering
    \includegraphics[width=.8\linewidth]{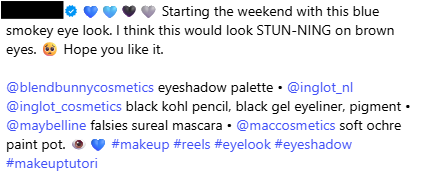}
    \caption{Example of an Instagram post that may be sponsored due to the presence of tagged brands.}
    \label{fig: Example of ads}
\end{figure}

Distinguishing ads from organic posts can be ambiguous (Figure~\ref{fig: Example of ads}); tagged brands may signal sponsorship or merely personal preference. Even with close scrutiny, regulators can misjudge cases, risking unfair penalties for legitimate influencers and causing complaints, as seen in~\citep{reclamecodenl_statement_2023} before the Dutch self-regulatory body \textit{Stichting Reclame Code (SRC)}\footnote{https://www.reclamecode.nl/over-de-src/over-de-src/}, where an independent jury justified its decision using legal reasoning.

The lack of transparency in influencer marketing is the largest issue consistently identified by self-regulatory bodies~\citep{noauthor_certification_2025, practice_influencer_2025, almed_monitoring_2024}. Self-regulators are industry organisations that make private rules for businesses. The main challenge for such bodies trying to measure compliance with their own rules in practice is the sheer amount of social media posts that can potentially contain commercial content. The fact that social media platforms do not allow anyone to thoroughly search their databases further complicates the enforcement of transparency standards. For practitioners, separating organic content from ads is the first step in assessing the compliance of influencer marketing with advertising law and self-regulatory codes. This is a laborious process that requires experts to spend their time viewing social media posts that might not contain any advertising. Commercially available software platforms aid in this process by using keyword filters\footnote{For example, https://www.influencermonitor.com/}, which are usually not accurate enough to eliminate all organic posts from a sample. 


In response to these challenges, computational research has sought to automate the detection of undisclosed ads~\citep{zarei_characterising_2020,kim_discovering_2021,martins_characterizing_2022,mathur_endorsements_2018,bertaglia2023closing,bertaglia2024instasynth}, but current methods face two limitations: (1)~they often lack a solid legal foundation, exposing regulators to pushback in relation to their decisions, and (2)~they prioritise accuracy over explanation~\citep{rogers_closed_2023}. Without a reasoning process, detection systems risk crossing boundaries that may conflict with free speech or other protected interests~\citep{huang_content_2025}. 

Large Language Models (LLMs) may offer promising ways to address both of these gaps. They can be prompted to reference relevant legal rules and provide explanations for their outputs, which makes the results more transparent and easier to interpret~\citep{louis_interpretable_2024}. At the same time, LLMs are prone to errors (e.g. hallucinations, weak grounding~\citep{dahl_large_2024,bang_hallulens_2025}). This paper explores the potential of LLMs for detecting undisclosed influencer marketing by examining how they identify hidden advertising and evaluating the quality of their accompanying legal reasoning. Our main contributions are (1)~a taxonomy of common errors in legal reasoning generated by LLM, extending previous research to a complex domain-specific context, namely, the detection of undisclosed advertisements on social networks, which also serves as a broader example of automatic compliance monitoring; (2)~an original dataset of LLM explanations annotated by two students who were trained in influencer marketing law; (3)~a demonstration of quantitative and qualitative evaluation strategies for LLM explanations and critically reflects on how these findings can support advertising regulatory bodies in automating moderation processes on a solid legal foundation. In general, our multidisciplinary approach, combining legal expertise with computer science, advances research on sponsored content detection and offers practical insights directly applicable to the industry. Finally, we make our material (data, code, annotation results) available.\footnote{https://github.com/HaoyangGui/Evaluating-LLM-Generated-Legal-Explanations}

\section{Related Work}
\subsection{NLP and sponsored content detection}


Recent advances in detecting hidden advertisements on social media leverage both rule-based and machine learning approaches~\citep{gui_computational_2025,bertaglia2025influencer}. Rule-based methods detect explicit cues such as coupon codes or campaign hashtags like `\#ad' with high precision~\citep{santos_rodrigues_identifying_2021,swart_is_2020}, but struggle with implicit or unconventional disclosures. Machine learning methods, in contrast, capture complex, context-dependent patterns from annotated datasets. For example, \citet{kim_discovering_2021} combined textual, visual, and social network features to improve detection, \citet{zarei_characterising_2020} identified a notable share of undisclosed Instagram promotions, and \citet{kok-shun_leveraging_2025} used GPT-4o to detect sponsored YouTube segments with high accuracy. Despite these gains, a shared limitation is that such models largely operate as black boxes, producing accurate predictions without interpretable reasoning.

\subsection{LLM and legal texts} 

Parallel to advances in sponsored content detection, research has explored the ability of LLMs to process legal text across tasks such as legal question answering~\citep{yuan_bringing_2024}, judgment prediction~\citep{medvedeva_legal_2023,chalkidis_lexglue_2022}, contract review~\citep{hendrycks_cuad_2021}, and legal reasoning~\citep{guha_legalbench_2023}, with reviews summarising tasks, datasets, methods, and challenges~\citep{katz_natural_2023,ariai_natural_2025}. General-purpose LLMs like GPT-4 and Claude perform well only after fine-tuning on legal examples~\citep{blair-stanek_blt_2024}, motivating benchmarks that consolidate legal tasks into unified evaluation frameworks~\citep{guha_legalbench_2023,fei_lawbench_2024} or building more interpretable legal question answering models using a Retrieval-Augmented Generation (RAG) approach~\citep{louis_interpretable_2024}. Related work has also examined LLM reasoning in legal adjacent domains, where the study is not only on the legal text but its application on real-world, real-user data, such as policy interpretation~\citep{palka_make_2025,palla_policy-as-prompt_2025} and content moderation~\citep{kolla_llm-mod_2024}, highlighting both the potential and challenges of applying LLMs to specialised, rule-governed contexts similar to law.

\subsection{Evaluating LLM output with legal knowledge}

Recent research has moved beyond measuring the raw accuracy of LLMs in legal and policy-related tasks to evaluating the trustworthiness and quality of their explanations~\citep{huang_content_2025, zhang_evaluation_2024,calderon_alternative_2025}. Across domains such as legal reasoning~\citep{kang_automating_2025,mishra_investigating_2025}, policy enforcement~\citep{palka_make_2025}, and content moderation~\citep{kolla_llm-mod_2024}, a key challenge is how to systematically assess LLM outputs in relation to legal knowledge. Despite this growing interest, progress is hindered by the lack of specific datasets that provide legally-informed annotations, which are critical for accurate benchmarking and systematically assessing both classification performance and the quality of generated legal reasoning.

Traditional evaluation metrics, such as accuracy, F1 score, and correlation, provide a baseline to assess classification performance~\citep{bavaresco_llms_2025,ashktorab_aligning_2025,tan_judgebench_2025,trautmann_measuring_2024}, but they fail to capture LLMs’ ability to understand context and nuance~\citep{huang_content_2025}. Some studies incorporate lexical and semantic similarity~\citep{vats_llms_2023}, while broader computational metrics examine conflict rates among LLM annotators~\citep{wang_large_2024}, plausibility and faithfulness of explanations~\citep{shailya_lext_2025}, groundedness~\citep{trautmann_measuring_2024}, and stability~\citep{blair-stanek_llms_2025}, often combined with statistical agreement with human experts~\citep{chiang_can_2023,calderon_alternative_2025}. 

Recognising that neither automated nor human judgments are perfectly accurate, recent work emphasises transparency in LLM-generated output, assessing qualities such as consistency, coherence, and informational richness~\citep{golovneva_roscoe_2023,prasad_receval_2023,patel_multi-logieval_2024}, alongside manually identifying reasoning errors~\citep{li_making_2023,tyen_llms_2024,mishra_investigating_2025}. For instance, \citet{mishra_investigating_2025} develops an error taxonomy for legal reasoning and methods to automate error detection. Collectively, these studies highlight that while LLMs show promise for legal and self-regulatory tasks, their out-of-the-box performance is limited, and fine-tuning is often required. Crucially, prior research has not extended these evaluation frameworks to complex, domain-specific contexts, such as legal interpretation in detecting undisclosed advertisements on social media, which is a key gap in compliance detection.

\section{Study design and methodology}

In this study, we evaluate how different LLMs classify influencer content and produce legal reasoning to justify their identification of advertising in the Dutch context. To this end, we created a dataset consisting of three types of content: disclosed advertisements, hidden advertisements, and organic posts (details are provided in the following section). The dataset is first fed into three different LLMs under three prompting strategies. Each model produces two outputs: (1)~a binary classification indicating whether the post constitutes an advertisement, and (2)~an accompanying explanation with legal reasoning to justify the decision. Then, for all posts and each type of content, we use two methods to examine the outputs:

\textbf{Quantitative evaluation}: We assess advertisement/organic content classification performance using standard classification metrics. This enables performance comparisons both within and across categories, and allows us to select the two best-performing models to proceed to the next step. Limiting further evaluation to these top-performing models helps avoid redundant comparisons and streamlines the analysis process. As a baseline, we use a TF-IDF (unigrams, bigrams) representation combined with logistic regression, employing an 80:20 train-test split. We did not include other deep learning models, such as BERT, as prior work suggests that they perform even worse in this context~\citep{bertaglia2023closing}.

\textbf{Qualitative evaluation}: We manually select balanced representative cases from each content type. Research assistants review the explanations by rating their helpfulness and annotating error types. This reveals systematic patterns linking specific errors to content types and prompting strategies. We also provide a case analysis, where a senior legal researcher reflects on the textual quality of a selection of outputs.

\subsection{Dataset}
\label{sec:Dataset}

\begin{table*}[ht]
\centering
\begin{tabular}{lccc}
\toprule
 & Disclosed & Organic & Undisclosed \\
\midrule
Posts & 592 & 424 & 127 \\
Tokens (mean $\pm$ std) & 51.67 $\pm$ 47.93 & 26.45 $\pm$ 45.65 & 41.04 $\pm$ 56.97 \\
Hashtags (mean $\pm$ std) & 2.23 $\pm$ 3.53 & 2.91 $\pm$ 7.18 & 1.84 $\pm$ 4.11 \\
Mentions (mean $\pm$ std) & 1.36 $\pm$ 1.06 & 0.51 $\pm$ 2.38 & 1.90 $\pm$ 2.19 \\
Posts with hashtag (\%) & 58.78 & 33.25 & 41.73 \\
Posts with mention (\%) & 90.88 & 12.74 & 93.70 \\
\bottomrule
\end{tabular}
\caption{Descriptive statistics for posts by category. Means and standard deviations (std) are reported for tokens, hashtags, and mentions. Posts with hashtags/mentions (\%) show the percentage of posts that have hashtags or mentions.}
\label{tab:post_stats}
\end{table*}

The dataset used in this study originates from~\citet{gui_across_2024} and comprises 300,199 posts by influencers registered in the Dutch Video-Uploader Registry~\footnote{https://www.cvdm.nl/registers/}. For the purposes of this research, we focus exclusively on Instagram as the platform of interest and restrict our analysis to posts written in English. In line with the standards established by~\citet{gui_across_2024}, we adopt the same criteria for identifying sponsorship disclosures. Specifically, we only include posts with so-called `green disclosures' (legally sufficient disclosed advertisement), which meet the legal requirements set out in the Dutch Advertising Code, resulting in 592 posts.

To construct a dataset for classification purposes, we then randomly sample an equal number of posts without green disclosures drawn from the same set of influencers (15 or the maximum number of posts by each), resulting in 551 posts. These posts may contain either sponsored content or not; therefore, three domain experts annotated these posts, distinguishing between hidden advertisements and organic content. The final labels are assigned through a two-step process: two domain experts (ann1 and ann2) must reach consensus, with any disagreements or uncertain cases referred to the third domain expert (ann3). Excluding 10.34\% uncertain cases, annotators 1 and 2 achieve a 92.64\% absolute agreement rate and 0.74 Krippendorff's Alpha, indicating substantive agreement.

The final dataset includes 1,143 English-language posts: 592 disclosed ads, 127 undisclosed ads, and 424 organic posts. To evaluate the ability of LLMs to detect hidden advertising, all explicit disclosure cues (such as \#ad, etc.) are removed from the disclosed ads before model input. Table~\ref{tab:post_stats} provides a detailed description of the dataset, showing that organic posts tend to be shorter and include fewer hashtags and mentions. In contrast, sponsored posts are generally more similar to each other than to organic content, which increases the challenge of accurately distinguishing between these categories.

\subsection{Models and prompts}
We employ three prompting strategies, each with identical task instructions but varying in the degree of provided legal knowledge. By gradually reducing the amount of legal context, we aim to examine the extent to which LLMs rely on and apply legal knowledge when identifying advertisements. In all cases, each prompt instructs the LLM to determine whether a post is advertising and to provide a legal reasoning explanation. The three levels of legal knowledge are defined as follows:
\begin{itemize}
    \item \textbf{Original codes with explanations}: This prompt incorporates the full regulatory text issued by \textit{Stichting Reclame Code (SRC)}, a Dutch self-regulatory organisation that promotes responsible advertising in addition to legislation. This prompt includes the original regulation text and the corresponding explanations from the \textit{General Section} and the special \textit{Advertising Code Social Media \& Influencer Marketing (RSM)}. This context is the most comprehensive form of legal knowledge based on text.
    \item \textbf{Original codes without explanations}: This prompt contains the same full regulatory text from the SRC as above, but omits the explanatory notes.
    \item \textbf{Names of the advertising codes only}: This prompt merely references the titles of the two codes (General Section and RSM), without including the substantive legal texts.
\end{itemize}

To ensure comparability, we designed a single base instruction prompt (shown in Appendix~\ref{appendix: base prompt}), which was adapted for each strategy. This base prompt was validated and refined through manual inspection of sample cases and iterative discussions among the co-authors. Although this process resulted in minor differences in wording across the three strategies, the overall task structure and requirements remained consistent. 

We evaluated the three prompting strategies using three different LLMs: \textit{gemini-2.5-flash-lite, gpt-4.1-nano, gpt-5-nano}. We ran all experiments with a temperature setting of 1 and used default values for all remaining hyperparameters.

\begin{table*}[ht]
\centering
\begin{tabular}{llrrr}
\toprule
Model & Prompting strategy & Precision & Recall & F1 \\
\midrule
logistic regression (TF-IDF)& & 0.85  &  0.91   &   0.88 \\
gemini-2.5-flash-lite & no\_article          & 0.91 & \textbf{0.93} & 0.92 \\
gemini-2.5-flash-lite & article             & 0.92 & \textbf{0.93} & \textbf{0.93} \\
gemini-2.5-flash-lite & article\_explanation & 0.92 & 0.92 & 0.92 \\
gpt-4.1-nano         & no\_article          & 0.88 & 0.87 & 0.87 \\
gpt-4.1-nano         & article             & 0.87 & 0.83 & 0.85 \\
gpt-4.1-nano         & article\_explanation & 0.86 & 0.83 & 0.85 \\
gpt-5-nano            & no\_article          & 0.94 & 0.91 & 0.92 \\
gpt-5-nano            & article             & 0.94 & 0.87 & 0.91 \\
gpt-5-nano            & article\_explanation & \textbf{0.95} & 0.86 & 0.90 \\
\bottomrule
\end{tabular}
\caption{Comparison of performance across models and prompting strategies for the whole dataset in the task of advertisement identification.}
\label{tab: whole dataset comparison}
\end{table*}

\subsection{Explanation evaluation: error annotations}
\label{sec:Explanation analysis: error annotations}
One of the objectives of this study is to examine the extent to which LLMs can comprehend legal knowledge and apply it to justify their decisions through legal reasoning. To assess the quality of the explanations produced by the models, we define seven common error categories: (e1) \textit{Wrong interpretation of legal citations}, (e2) \textit{No citation}, (e3) \textit{Citation is not clear}, (e4) \textit{Hallucinations on the legal citations}, (e5) \textit{Hallucinations on the content}, (e6) \textit{Mistaken potential cues}, and (e7) \textit{Reasoning results in opposite output}. Detailed descriptions and examples are provided in Table \ref{tab:llm_errors} (Appendix~\ref{appendix:unified}).

Two research assistants with legal knowledge (annA and annB) rated the helpfulness of a subset of explanations and annotated the presence of these errors. Since LLM outputs vary widely in length and content, we only note whether an error is present in an explanation (note that one explanation might contain multiple errors). Before annotation, the assistants received training from domain experts and completed revisions after resolving any ambiguities.

The evaluation sample includes 60 randomly selected posts, evenly distributed across three types of content: 20 disclosed ads, 20 hidden ads, and 20 organic posts. For hidden ads and organic posts, we further divide the 20 examples into two groups based on the earlier sponsorship annotation stage: 10 posts with consensus labels from annotators ann1 and ann2, and 10 labelled solely by ann3 (no consensus reached by ann1 and ann2).

For the evaluation of the explanations, annotators A and B labelled 10 overlapping posts (in addition to 25 distinct posts each), achieving 89.29\% absolute agreement and 0.37 Krippendorf's Alpha. As we compare different LLMs (\textit{gpt-5-nano} and \textit{gemini-2.5-flash-lite}) under three prompting strategies, each annotator evaluates 210 explanation units (35 posts $\times$ 2 models $\times$ 3 prompting strategies). Using these annotations, we analyse and discuss how explanation quality varies across models, prompting strategies, and different types of content.

\subsection{Explanation evaluation: case analysis}
We complement the evaluation of the explanations with a qualitative, expert-driven evaluation of the results. For this, one of the authors of this paper, a senior legal scholar with expertise in Dutch advertising law, was assigned a random set of four explanations pertaining to two posts from the article\_explanation prompt, one of which involves a disclosed advertisement and the other an undisclosed advertisement. While these examples cannot capture every factor present in the dataset, this case analysis provides insight into the recurring patterns that characterise each experimental setting.

\label{sec:Explanation analysis: case analysis}

\section{Results}

We first evaluate the classification performance of three LLMs under three prompting strategies across the entire dataset in a zero-shot setup (i.e., without fine-tuning). Based on these results, we select the two best-performing LLMs for subsequent tasks, which include evaluating classification performance on each type of content and examining the quality of their explanations. 
\subsection{Classification results}

Table~\ref{tab: whole dataset comparison} presents classification performance on the full dataset of 1,143 posts across all experimental settings. Overall, the results indicate that all models achieve reasonable performance, but \textit{gpt-4.1-nano} consistently underperforms on every metric, even worse than the baseline, with F1 scores ranging from 0.85 to 0.87. To streamline further analyses, we focus on \textit{gpt-5-nano} (GPT) and \textit{gemini-2.5-flash-lite} (Gemini).

Examining model-level performance, GPT achieves the highest precision (0.95 with the article\_explanation prompt), while Gemini demonstrates stronger recall (0.93) and generally higher F1 scores (0.93). Interestingly, the prompting strategy that incorporates the most legal knowledge (article\_explanation) does not always yield the best overall classification performance. For GPT, although article\_explanation maximises precision, it reduces recall, resulting in the lowest F1 (0.90). Similarly, for Gemini, the highest recall (0.93) is achieved without explanations (article prompt), highlighting that more legal knowledge does not automatically translate into better classification outcomes. Differences across prompting strategies are relatively small, but this pattern suggests that LLMs’ ability to apply legal knowledge may rely more on patterns learned during pretraining rather than the provided legal text.

Next, we focused on 95 ambiguous posts where annotators (ann1 and ann2) disagreed or expressed uncertainty in the advertisement annotation procedure (section~\ref{sec:Dataset}). As expected, overall performance dropped significantly, with F1 scores falling by over 10 percentage points compared to the full dataset. The baseline model exhibited an even steeper decline, exceeding a 30-point reduction. GPT shows high precision (0.80 with no\_article prompt) but suffers from lower recall, whereas Gemini maintains stronger recall and balanced F1 scores (0.80), consistent with its relative strengths in the full dataset. Notably, no prompting strategy equipped with explanations consistently outperforms others, reinforcing the observation that adding explicit legal text does not guarantee improved performance, particularly on ambiguous or borderline cases. Detailed results are provided in Table~\ref{tab:performance for tricky cases} (Appendix~\ref{appendix:unified}).

Zooming in on the results by types of content, Gemini performs better on disclosed and hidden ads (0.94 and 0.93), whereas GPT performs better on organic content (0.92). GPT’s performance on hidden ads remains notably weaker, even weaker than the baseline model, suggesting that its precision-oriented strengths do not extend to detecting subtle or undisclosed advertising cues. Prompting strategies show no consistent pattern: for Gemini, `article' prompts perform best overall, while `no\_article' prompts slightly lead on disclosed and hidden ads; for GPT, `no\_article' prompts dominate on disclosed and hidden ads, whereas legal-knowledge prompts are better for organic content. A more granular breakdown of accuracy by content type, model, and prompting strategy can be found in Table \ref{tab:content_category_accuracy} (Appendix~\ref{appendix:unified}).


\subsection{Evaluation of explanations}
To assess the quality of LLM-generated legal explanations, we consider two complementary dimensions: (1)~their perceived helpfulness to annotators, and (2)~the types and frequencies of errors they contain.
\begin{table*}[ht]
\centering
\resizebox{\textwidth}{!}{%
\begin{tabular}{llllrrrrrrr}
\toprule
 & Model & Variant & Helpfulness score & e1 (\%) & e2 (\%) & e3 (\%) & e4 (\%) & e5 (\%) & e6 (\%) & e7 (\%) \\
\midrule
0 & gemini-2.5-flash-lite & no\_article & 3.31 $\pm$ 0.84 & \textbf{17.14} & \textbf{81.43} & 25.71 & 5.71 & 2.86 & 10.00 & 0.00 \\
1 & gemini-2.5-flash-lite & article & \textbf{4.37 $\pm$ 0.89 }& 5.71 & 5.71 & 7.14 & 0.00 &\textbf{ 7.14} & \textbf{12.86} & 1.43 \\
2 & gemini-2.5-flash-lite & article\_explanation & \textbf{4.37 $\pm$ 0.75} & 7.14 & 2.86 & 4.29 & 0.00 & 1.43 & 10.00 & 0.00 \\
3 & gpt-5-nano & no\_article & 3.29 $\pm$ 0.93 & 2.86 & 78.57 & \textbf{35.71} & \textbf{7.14} & 1.43 & 2.86 & 0.00 \\
4 & gpt-5-nano & article & 4.20 $\pm$ 1.10 & 11.43 & 1.43 & 20.00 & 1.43 & 1.43 & 4.29 & 0.00 \\
5 & gpt-5-nano & article\_explanation & 4.13 $\pm$ 0.99 & 7.14 & 1.43 & 31.43 & 1.43 & 0.00 & 4.29 & 0.00 \\
Total &  &  &  & 8.57 & 28.57 & 20.71 & 2.62 & 2.38 & 7.38 & 0.24 \\
\bottomrule
\end{tabular}
}
\caption{Helpfulness score and error rates across models and prompting strategies for all types of content. Error types: (e1) \textit{Wrong interpretation of legal citations}, (e2) \textit{No citation}, (e3) \textit{Citation is not clear}, (e4) \textit{Hallucinations on the legal citations}, (e5) \textit{Hallucinations on the content}, (e6) \textit{Mistaken potential cues}, and (e7) \textit{Reasoning results in opposite output}. For each error, the value shown is the proportion of posts containing the corresponding error. The last row shows the percentage of each error across the whole dataset.}

\label{tab:helpfulness_errors}
\end{table*}
\paragraph{Helpfulness and errors by models and prompting strategies}

We begin by analysing the errors in the explanations as described above. The last row in Table~\ref{tab:helpfulness_errors} shows the percentage of error types observed in LLM-generated explanations across all annotated posts. The most frequent error is e2 (No citation, 28.57\%), followed by e3 (Unclear citation, 20.71\%), indicating that LLMs often attempt but fail to provide explicit legal references. Less common errors include e1 (Wrong interpretation, 8.57\%), e6 (Mistaken cues, 7.38\%), e4 (Hallucinated citations, 2.62\%), and e5 (Hallucinated content, 2.38\%), while e7 (Contradictory reasoning, 0.24\%) is rare. These patterns raise a key question: do models genuinely understand legal content or simply produce superficially plausible explanations?


Table~\ref{tab:helpfulness_errors} also presents a detailed assessment of LLM explanations in terms of both their perceived helpfulness and the percentage of data that contains corresponding types of errors across different models and prompting strategies. Helpfulness scores (1–5 scale) show Gemini with article\_explanation performs best (4.37 $\pm$ 0.75), followed by GPT with article prompts (4.20 $\pm$ 1.10). No\_article variants for both models achieve the lowest scores, indicating that legal input, especially when combined with explanations, improves perceived reasoning quality.



Critical citation errors (e2, e3) dominate no\_article prompts: 81.43\% and 25.71\% for Gemini, 78.57\% and 35.71\% for GPT. Even with legal input, GPT still shows notable e1/e3 rates (7.14\%/31.43\% for article\_explanation, 11.43\%/20\% for article), whereas Gemini’s rates are lower (7.14\%/4.29\% for article\_explanation, 5.71\%/7.14\% for article). However, Gemini exhibits higher e6 under article prompts (12.86\%), showing that legal text alone does not guarantee accurate interpretation. In contrast, hallucinations (e4, e5) remain rare but concerning. Nearly all e4 cases occur in no\_article prompts (5.71\% Gemini, 7.14\% GPT), where models fabricate citations due to missing legal context.


\paragraph{Errors by content type}


We also analyse errors by content type. Disclosed ads show the lowest rates for most errors, except for some e1–e3 cases in no\_article variants. With legal context, hallucinations (e4, e5) are virtually absent, indicating that models rarely fabricate legal citations or misrepresent content when sufficient context is given. Detailed results can be found in Table~\ref{tab:errors by content} (Appendix~\ref{appendix:unified}).


Undisclosed ads exhibit the highest e6 rate (28.57\%) and notable e3 errors, with e4 and e5 appearing more often than in other categories. These patterns reflect the difficulty of detecting subtle promotions, where models must infer intent from indirect cues and often misidentify which signals indicate sponsorship.


Organic content shows comparatively higher e2 (No citation) and e3 (Unclear citation) errors, especially under no\_article prompts, suggesting that models sometimes false legal reasoning without a real basis. Moderate e6 levels further indicate a tendency to overfit and misread ordinary content as promotional, highlighting the inherent ambiguity of influencer posts.

\paragraph{Case analysis: examining legal reasoning}
From a legal perspective, the task is simple, albeit domain-specific. Legal explanations follow an innate structure, due to the relevance of logic for legal argumentation~\citep{bench-capon_argumentation_2009,lind_significance_2014}. The task at hand involves identifying whether a post constitutes advertising. Our case analysis reveals that neither model was able to generate a cohesive, well-structured legal explanation. The model outputs an amalgam of statements, which is comparable to a rather poorly performing first-year law student. To be considered a basic but complete legal analysis, the output needed better performance in terms of selecting relevant provisions and in terms of structure. 

In terms of provisions, according to the Dutch Advertising Code, which is industry self-regulation in the Netherlands, the starting point in determining whether something is advertising is that it has to fulfil all the conditions of Article 1 in~\citet{stichting_reclame_code_general_nodate} and Article 2. (c, d, e) in~\citet{stichting_reclame_code_special_2023}.
While some dimensions of this definition cannot be analysed without additional facts (e.g., the relationship between an advertiser and a third party), some very concrete conditions should have been considered in an explanation, such as whether a post on Instagram is public, whether the promotion of goods or services is direct or indirect, or whether the post consists in an idea, a good or a service. The four explanations in our case analysis mention Article 1, but there is generally a lack of systematic tackling of the conditions. In addition, the models seem to try to select and discuss many other articles, sometimes irrelevant (e.g., GPT mentioning Article 8.4). In terms of structure, there is no acknowledgement that a legal analysis is a demonstration that needs to be built according to some form of structure. 

Generally, such a structure will differ from country to country or across fields of legal theory and practice; an inherent and easily detectable logic is necessary. All four explanations seem to provide some sort of conclusion, whether explicitly recognised as such or not, but the conclusion sometimes makes logical jumps, or it is a demonstration of conditions which are not relevant.   
Based on these factors, the explanations might seem, at first sight, to have relevance and accuracy, but upon closer examination, they are either chaotic, incomplete, or simply inaccurate. 

\section{Discussion and conclusion}
This study examined how large language models (LLMs) can be applied to detect undisclosed advertising on social media while providing legal reasoning. Unlike prior research, which focused almost exclusively on classification accuracy, our work systematically evaluates both the quality of classification and the legal soundness of LLM explanations. This dual lens highlights critical gaps in current practice and suggests pathways toward more transparent and accountable automated moderation systems.

Starting from the classification task, both \textit{gpt-5-nano} and \textit{gemini-2.5-flash-lite} achieve high overall accuracy in identifying advertising content, but model choice strongly influences both classification strength and error profile: Gemini is more effective for recall-oriented tasks such as detecting hidden ads, whereas GPT excels in precision. Notably, LLMs are not always superior to simple baselines in overall classification performance; however, they perform better in challenging cases. Similar patterns of strength appear in the 95 ambiguous posts, where annotators (ann1 and ann2) disagreed or expressed uncertainty in the advertisement annotation procedure (section 3.1). Examining the content further, these patterns of ambiguity align with previous findings, which attribute annotator disagreement to both data-related factors (e.g., various language features, uncertainty in sentence meaning), and annotator-related factors (e.g. various language features, uncertainty in sentence meaning)~\citet{jiang_investigating_2022, plank_problem_2022, xu_dissonance_2023}. These intrinsic complexities pose challenges for LLMs, contributing to lower performance in ambiguous contexts.

Moreover, increasing the amount of embedded legal text does not consistently improve the classification outcomes. While prompts containing full regulatory codes and explanations raise the perceived helpfulness of LLM reasoning (e.g., Gemini article\_explanation reaching 4.37 $\pm$ 0.75 versus 4.20 $\pm$ 1.10 for GPT), they do not guarantee better moderation outcomes. This indicates that current LLMs do not simply ``read and apply'' legal norms; instead, they rely heavily on internal heuristics and contextual associations. In practice, this means that LLMs are already capable of recognising different forms of advertising because promotional language and stylistic cues are strongly represented in their training data. Cues indicating sponsorship, patterns of product placement, or persuasive rhetorical devices can often be detected without direct reference to regulatory codes. In this sense, the models’ performance may reflect an underlying competence in identifying pragmatic markers of advertising, rather than understanding and applying legal knowledge as a content moderator.

The explanation analysis further reveals systematic weaknesses. Citation-related errors, missing (e2, 28.57\%), unclear (e3, 20.71\%), or wrong interpretations (e1, 8.57\%), dominate across settings, particularly when no legal text is provided. Even when legal sources are available, models often select irrelevant provisions or fail to structure reasoning in a way consistent with basic legal methodology. More severe hallucinations of legal citations (e4, 2.62\%) and content (e5, 2.38\%) are rare but concentrated in no\_article prompts, where GPT and Gemini fabricated legal references at 7.14\% and 5.71\%, respectively. These patterns suggest that LLMs tend to approximate legal reasoning rather than reliably apply normative rules, which essentially means that they fail to 'read, understand, and apply.'

A closer look by content type further illuminates these limitations. Undisclosed ads produce the highest rate of misidentified cues (e6, 28.57\%), showing that LLMs frequently mistake ordinary or ambiguous content for sponsored posts. In contrast, disclosed ads show almost no hallucinations when legal text is provided, indicating that straightforward content allows LLMs to stabilise their reasoning more reliably. Together with the case analysis carried out, these findings confirm that although LLMs can approximate legal reasoning, they are far from delivering rigorous justifications akin to an expert with domain-specific knowledge.

These findings have two broad implications for moderation. First, they demonstrate that high classification accuracy does not ensure trustworthy enforcement. An LLM that labels a post correctly but cites irrelevant or fabricated legal provisions cannot satisfy procedural fairness standards. Second, explanation quality varies systematically by content type and prompting strategy, meaning that moderation pipelines cannot rely on a one-size-fits-all approach. Platforms using LLMs for detection must pair performance metrics with legal-reasoning audits to ensure that decisions are not only correct but also defensible. In practice, this means building tools that flag cases with high-risk errors (e.g., e4/e5 hallucinations) for human review and calibrating models to reduce over-classification in ambiguous contexts.

These findings also connect to broader debates on moderation with LLMs. As \citet{goanta_regulation_2023} argues, NLP research must be situated within regulatory studies to avoid regulatory capture and to bridge the ``pacing gap'' between technological innovation and legal adaptation. Our results reflect this concern: models that appear accurate can still misapply or fabricate legal norms, undermining the legitimacy of enforcement. Treating moderation as a purely technical task risks obscuring the regulatory standards it is supposed to serve; instead, explanation quality and legal soundness must be foregrounded alongside accuracy. At the same time, our taxonomy of explanation errors resonates with emerging moderation research that highlights the concerns of LLMs as moderators. \citet{yin_bingoguard_2025} demonstrates that binary safe/unsafe labels miss important gradations of harm. Similarly, in our research, not all explanation errors are equally harmful: vague reasoning may be tolerable, but fabricated citations or misapplied provisions threaten procedural fairness. Integrating severity-sensitive auditing into compliance monitoring would thus allow regulators to triage high-risk cases while ensuring that enforcement remains both effective and legitimate.

The main contribution of this paper is to integrate the quality of legal reasoning in the evaluation of influencer marketing detection systems. By developing a taxonomy of LLM explanation errors and showing how these patterns vary by model, prompting strategy, and content type, we provide an actionable framework for regulators and platform designers. Instead of treating LLM outputs as opaque predictions, our study demonstrates how to assess whether automated moderation is not only accurate but also legitimate. This is particularly valuable for self-regulatory bodies such as \textit{Stichting Reclame Code (SRC)}, which must justify enforcement decisions in legal terms rather than through statistical metrics alone. More broadly, our multidisciplinary approach, combining computational evaluation with legal analysis, offers a blueprint for building moderation systems that are transparent, explainable, and aligned with rule-of-law principles rather than black-box heuristics.

\section*{Limitations}
Our dataset focuses solely on textual content, excluding visual or multimodal signals that frequently convey sponsorship. Human annotation also entails subjectivity, especially for borderline cases where even experts disagree. Moreover, the study relies on off-the-shelf LLMs without fine-tuning, meaning performance could improve with domain-specific adaptation.

\section*{Acknowledgments}
This research has been supported by funding from the ERC Starting Grant HUMANads (ERC-2021-StG No 101041824). We also thank Isolde Torres and Giulio Bernasconi for their valuable assistance with this research.

\bibliography{custom}

\begin{thebibliography}{62}
\providecommand{\natexlab}[1]{#1}

\bibitem[{Almed(2024)}]{almed_monitoring_2024}
{IAP} Almed. 2024.
\newblock \href {chrome-extension://efaidnbmnnnibpcajpcglclefindmkaj/https://www.iap.it/wp-content/uploads/2025/04/Monitoring-Transparency-and-Influencer-Marketing-Report-2024-IAP-Almed.pdf} {Monitoring {{Transparency}} and {{Influencer Marketing}}: Beauty, fashion, family and finance}.

\bibitem[{Ariai and Demartini(2025)}]{ariai_natural_2025}
Farid Ariai and Gianluca Demartini. 2025.
\newblock \href {https://doi.org/10.48550/arXiv.2410.21306} {Natural language processing for the legal domain: A survey of tasks, datasets, models, and challenges}.
\newblock \emph{Preprint}, arXiv:2410.21306.

\bibitem[{Ashktorab et~al.(2025)Ashktorab, Desmond, Pan, Johnson, Cooper, Daly, Nair, Pedapati, Do, and Geyer}]{ashktorab_aligning_2025}
Zahra Ashktorab, Michael Desmond, Qian Pan, James~M. Johnson, Martin~Santillan Cooper, Elizabeth~M. Daly, Rahul Nair, Tejaswini Pedapati, Hyo~Jin Do, and Werner Geyer. 2025.
\newblock \href {https://doi.org/10.48550/arXiv.2410.00873} {Aligning human and {{LLM}} judgments: {{Insights}} from {{EvalAssist}} on task-specific evaluations and {{AI-assisted}} assessment strategy preferences}.
\newblock \emph{Preprint}, arXiv:2410.00873.

\bibitem[{Bang et~al.(2025)Bang, Ji, Schelten, Hartshorn, Fowler, Zhang, Cancedda, and Fung}]{bang_hallulens_2025}
Yejin Bang, Ziwei Ji, Alan Schelten, Anthony Hartshorn, Tara Fowler, Cheng Zhang, Nicola Cancedda, and Pascale Fung. 2025.
\newblock \href {https://doi.org/10.48550/arXiv.2504.17550} {{{HalluLens}}: {{LLM}} hallucination benchmark}.
\newblock \emph{Preprint}, arXiv:2504.17550.

\bibitem[{Bavaresco et~al.(2025)Bavaresco, Bernardi, Bertolazzi, Elliott, Fernández, Gatt, Ghaleb, Giulianelli, Hanna, Koller, Martins, Mondorf, Neplenbroek, Pezzelle, Plank, Schlangen, Suglia, Surikuchi, Takmaz, and Testoni}]{bavaresco_llms_2025}
Anna Bavaresco, Raffaella Bernardi, Leonardo Bertolazzi, Desmond Elliott, Raquel Fernández, Albert Gatt, Esam Ghaleb, Mario Giulianelli, Michael Hanna, Alexander Koller, Andre Martins, Philipp Mondorf, Vera Neplenbroek, Sandro Pezzelle, Barbara Plank, David Schlangen, Alessandro Suglia, Aditya~K Surikuchi, Ece Takmaz, and Alberto Testoni. 2025.
\newblock \href {https://doi.org/10.18653/v1/2025.acl-short.20} {{{LLMs}} instead of human judges? {{A}} large scale empirical study across 20 {{NLP}} evaluation tasks}.
\newblock In \emph{Proceedings of the 63rd Annual Meeting of the Association for Computational Linguistics (Volume 2: {{Short}} Papers)}, pages 238--255. Association for Computational Linguistics.

\bibitem[{Bench-Capon et~al.(2009)Bench-Capon, Prakken, and Sartor}]{bench-capon_argumentation_2009}
Trevor Bench-Capon, Henry Prakken, and Giovanni Sartor. 2009.
\newblock \href {https://doi.org/10.1007/978-0-387-98197-0_18} {Argumentation in legal reasoning}.
\newblock In Guillermo Simari and Iyad Rahwan, editors, \emph{Argumentation in Artificial Intelligence}, pages 363--382. Springer US.

\bibitem[{Bertaglia et~al.(2025)Bertaglia, Goanta, Spanakis, and Iamnitchi}]{bertaglia2025influencer}
Thales Bertaglia, Catalina Goanta, Gerasimos Spanakis, and Adriana Iamnitchi. 2025.
\newblock Influencer self-disclosure practices on {{Instagram}}: {{A}} multi-country longitudinal study.
\newblock \emph{Online Social Networks and Media}, 45:100298.

\bibitem[{Bertaglia et~al.(2024)Bertaglia, Heisig, Kaushal, and Iamnitchi}]{bertaglia2024instasynth}
Thales Bertaglia, Lily Heisig, Rishabh Kaushal, and Adriana Iamnitchi. 2024.
\newblock Instasynth: {{Opportunities}} and challenges in generating synthetic instagram data with chatgpt for sponsored content detection.
\newblock In \emph{Proceedings of the International {{AAAI}} Conference on Web and Social Media}, volume~18, pages 139--151.

\bibitem[{Bertaglia et~al.(2023)Bertaglia, Huber, Goanta, Spanakis, and Iamnitchi}]{bertaglia2023closing}
Thales Bertaglia, Stefan Huber, Catalina Goanta, Gerasimos Spanakis, and Adriana Iamnitchi. 2023.
\newblock Closing the loop: {{Testing}} chatgpt to generate model explanations to improve human labelling of sponsored content on social media.
\newblock In \emph{World Conference on Explainable Artificial Intelligence}, pages 198--213. Springer.

\bibitem[{Blair-Stanek and Durme(2025)}]{blair-stanek_llms_2025}
Andrew Blair-Stanek and Benjamin~Van Durme. 2025.
\newblock \href {https://doi.org/10.48550/arXiv.2502.05196} {{{LLMs}} provide unstable answers to legal questions}.
\newblock \emph{Preprint}, arXiv:2502.05196.

\bibitem[{Blair-Stanek et~al.(2024)Blair-Stanek, Holzenberger, and Van~Durme}]{blair-stanek_blt_2024}
Andrew Blair-Stanek, Nils Holzenberger, and Benjamin Van~Durme. 2024.
\newblock \href {https://doi.org/10.18653/v1/2024.nllp-1.18} {{{BLT}}: {{Can}} large language models handle basic legal text?}
\newblock In \emph{Proceedings of the Natural Legal Language Processing Workshop 2024}, pages 216--232. Association for Computational Linguistics.

\bibitem[{Calderon et~al.(2025)Calderon, Reichart, and Dror}]{calderon_alternative_2025}
Nitay Calderon, Roi Reichart, and Rotem Dror. 2025.
\newblock \href {https://doi.org/10.48550/arXiv.2501.10970} {The alternative annotator test for {{LLM-as-a-judge}}: {{How}} to statistically justify replacing human annotators with {{LLMs}}}.
\newblock \emph{Preprint}, arXiv:2501.10970.

\bibitem[{Chalkidis et~al.(2022)Chalkidis, Jana, Hartung, Bommarito, Androutsopoulos, Katz, and Aletras}]{chalkidis_lexglue_2022}
Ilias Chalkidis, Abhik Jana, Dirk Hartung, Michael Bommarito, Ion Androutsopoulos, Daniel Katz, and Nikolaos Aletras. 2022.
\newblock \href {https://doi.org/10.18653/v1/2022.acl-long.297} {{{LexGLUE}}: A benchmark dataset for legal language understanding in english}.
\newblock In \emph{Proceedings of the 60th Annual Meeting of the Association for Computational Linguistics (Volume 1: {{Long}} Papers)}, pages 4310--4330. Association for Computational Linguistics.

\bibitem[{Chiang and Lee(2023)}]{chiang_can_2023}
Cheng-Han Chiang and Hung-yi Lee. 2023.
\newblock \href {https://doi.org/10.18653/v1/2023.acl-long.870} {Can large language models be an alternative to human evaluations?}
\newblock In \emph{Proceedings of the 61st Annual Meeting of the Association for Computational Linguistics (Volume 1: {{Long}} Papers)}, pages 15607--15631. Association for Computational Linguistics.

\bibitem[{Code(2023{\natexlab{a}})}]{stichting_reclame_code_general_nodate}
Stichting~Reclame Code. 2023{\natexlab{a}}.
\newblock \href {https://www.reclamecode.nl/engels/dutch-advertising-code/general/} {General - stichting reclame code}.

\bibitem[{Code(2023{\natexlab{b}})}]{stichting_reclame_code_special_2023}
Stichting~Reclame Code. 2023{\natexlab{b}}.
\newblock \href {https://www.reclamecode.nl/nederlandse-reclame-code/bijzondere-reclamecodes/#reclamecode-social-media---influencer-marketing--rsm-} {Special advertising codes - advertising code foundation}.

\bibitem[{Code(2023{\natexlab{c}})}]{reclamecodenl_statement_2023}
Stichting~Reclame Code. 2023{\natexlab{c}}.
\newblock \href {https://www.reclamecode.nl/uitspraak/?uitspraakId=447283} {Statement - advertising code foundation. {{Stichting Reclame Code}}}.

\bibitem[{Code(2025)}]{noauthor_certification_2025}
Stichting~Reclame Code. 2025.
\newblock \href {https://www.reclamecode.nl/nieuws-agenda/nieuws/certificering-werkt-overtredingen-influencers-halveren-na-e-learning/} {Certification works: Influencer violations halved after e-learning - {{Advertising Code Foundation}}}.

\bibitem[{Dahl et~al.(2024)Dahl, Magesh, Suzgun, and Ho}]{dahl_large_2024}
Matthew Dahl, Varun Magesh, Mirac Suzgun, and Daniel~E Ho. 2024.
\newblock \href {https://doi.org/10.1093/jla/laae003} {Large legal fictions: {{Profiling}} legal hallucinations in large language models}.
\newblock \emph{Journal of Legal Analysis}, 16(1):64--93.

\bibitem[{De~Veirman et~al.(2017)De~Veirman, Cauberghe, and Hudders}]{de_veirman_marketing_2017}
Marijke De~Veirman, Veroline Cauberghe, and Liselot Hudders. 2017.
\newblock \href {https://doi.org/10.1080/02650487.2017.1348035} {Marketing through {{Instagram}} influencers: The impact of number of followers and product divergence on brand attitude}.
\newblock \emph{International Journal of Advertising}, 36(5):798--828.

\bibitem[{Ershov and Mitchell(2020)}]{ershov_effects_2020}
Daniel Ershov and Matthew Mitchell. 2020.
\newblock \href {https://doi.org/10.1145/3391403.3399477} {The effects of influencer advertising disclosure regulations: {{Evidence}} from instagram}.
\newblock In \emph{Proceedings of the 21st {{ACM}} Conference on Economics and Computation}, {{EC}} '20, pages 73--74. Association for Computing Machinery.

\bibitem[{Fei et~al.(2024)Fei, Shen, Zhu, Zhou, Han, Huang, Zhang, Chen, Yin, Shen, Ge, and Ng}]{fei_lawbench_2024}
Zhiwei Fei, Xiaoyu Shen, Dawei Zhu, Fengzhe Zhou, Zhuo Han, Alan Huang, Songyang Zhang, Kai Chen, Zhixin Yin, Zongwen Shen, Jidong Ge, and Vincent Ng. 2024.
\newblock \href {https://doi.org/10.18653/v1/2024.emnlp-main.452} {{{LawBench}}: {{Benchmarking}} legal knowledge of large language models}.
\newblock In \emph{Proceedings of the 2024 Conference on Empirical Methods in Natural Language Processing}, pages 7933--7962. Association for Computational Linguistics.

\bibitem[{Goanta et~al.(2023)Goanta, Aletras, Chalkidis, Ranchordás, and Spanakis}]{goanta_regulation_2023}
Catalina Goanta, Nikolaos Aletras, Ilias Chalkidis, Sofia Ranchordás, and Gerasimos Spanakis. 2023.
\newblock \href {https://doi.org/10.18653/v1/2023.emnlp-main.539} {Regulation and {{NLP}} ({{RegNLP}}): {{Taming}} large language models}.
\newblock In \emph{Proceedings of the 2023 Conference on Empirical Methods in Natural Language Processing}, pages 8712--8724. Association for Computational Linguistics.

\bibitem[{Golovneva et~al.(2023)Golovneva, Chen, Poff, Corredor, Zettlemoyer, Fazel-Zarandi, and Celikyilmaz}]{golovneva_roscoe_2023}
Olga Golovneva, Moya Chen, Spencer Poff, Martin Corredor, Luke Zettlemoyer, Maryam Fazel-Zarandi, and Asli Celikyilmaz. 2023.
\newblock \href {https://doi.org/10.48550/arXiv.2212.07919} {{{ROSCOE}}: A suite of metrics for scoring step-by-step reasoning}.
\newblock \emph{Preprint}, arXiv:2212.07919.

\bibitem[{Guha et~al.(2023)Guha, Nyarko, Ho, Ré, Chilton, Narayana, Chohlas-Wood, Peters, Waldon, Rockmore, Zambrano, Talisman, Hoque, Surani, Fagan, Sarfaty, Dickinson, Porat, Hegland, Wu, Nudell, Niklaus, Nay, Choi, Tobia, Hagan, Ma, Livermore, Rasumov-Rahe, Holzenberger, Kolt, Henderson, Rehaag, Goel, Gao, Williams, Gandhi, Zur, Iyer, and Li}]{guha_legalbench_2023}
Neel Guha, Julian Nyarko, Daniel~E. Ho, Christopher Ré, Adam Chilton, Aditya Narayana, Alex Chohlas-Wood, Austin Peters, Brandon Waldon, Daniel~N. Rockmore, Diego Zambrano, Dmitry Talisman, Enam Hoque, Faiz Surani, Frank Fagan, Galit Sarfaty, Gregory~M. Dickinson, Haggai Porat, Jason Hegland, and 21 others. 2023.
\newblock \href {https://doi.org/10.48550/arXiv.2308.11462} {{{LegalBench}}: A collaboratively built benchmark for measuring legal reasoning in large language models}.
\newblock \emph{Preprint}, arXiv:2308.11462.

\bibitem[{Gui et~al.(2024)Gui, Bertaglia, Goanta, de~Vries, and Spanakis}]{gui_across_2024}
Haoyang Gui, Thales Bertaglia, Catalina Goanta, Sybe de~Vries, and Gerasimos Spanakis. 2024.
\newblock \href {https://doi.org/10.1007/978-3-031-78548-1_1} {Across platforms and languages: {{Dutch}} influencers and legal disclosures on instagram, {{YouTube}} and {{TikTok}}}.
\newblock In \emph{Social Networks Analysis and Mining: 16th International Conference, {{ASONAM}} 2024, Rende, Italy, September 2–5, 2024, Proceedings, Part {{III}}}, pages 3--12. Springer-Verlag.

\bibitem[{Gui et~al.(2025)Gui, Bertaglia, Goanta, and Spanakis}]{gui_computational_2025}
Haoyang Gui, Thales Bertaglia, Catalina Goanta, and Gerasimos Spanakis. 2025.
\newblock \href {https://doi.org/10.48550/arXiv.2506.14602} {Computational studies in influencer marketing: A systematic literature review}.
\newblock \emph{Preprint}, arXiv:2506.14602.

\bibitem[{Hendrycks et~al.(2021)Hendrycks, Burns, Chen, and Ball}]{hendrycks_cuad_2021}
Dan Hendrycks, Collin Burns, Anya Chen, and Spencer Ball. 2021.
\newblock \href {https://doi.org/10.48550/arXiv.2103.06268} {{{CUAD}}: {{An}} expert-annotated {{NLP}} dataset for legal contract review}.
\newblock \emph{Preprint}, arXiv:2103.06268.

\bibitem[{Huang(2025)}]{huang_content_2025}
Tao Huang. 2025.
\newblock \href {https://doi.org/10.48550/arXiv.2409.03219} {Content moderation by {{LLM}}: {{From}} accuracy to legitimacy}.
\newblock \emph{Preprint}, arXiv:2409.03219.

\bibitem[{Jiang and de~Marneffe(2022)}]{jiang_investigating_2022}
Nan-Jiang Jiang and Marie-Catherine de~Marneffe. 2022.
\newblock \href {https://doi.org/10.1162/tacl_a_00523} {Investigating reasons for disagreement in natural language inference}.
\newblock \emph{Transactions of the Association for Computational Linguistics}, 10:1357--1374.

\bibitem[{Kang et~al.(2025)Kang, Qu, Soon, Li, and Trakic}]{kang_automating_2025}
Xiaoxi Kang, Lizhen Qu, Lay-Ki Soon, Zhuang Li, and Adnan Trakic. 2025.
\newblock \href {https://doi.org/10.1007/s10506-025-09467-5} {Automating {{IRAC}} analysis in malaysian contract law using a semi-structured knowledge base}.
\newblock \emph{Artificial Intelligence and Law}.

\bibitem[{Katz et~al.(2023)Katz, Hartung, Gerlach, Jana, and II}]{katz_natural_2023}
Daniel~Martin Katz, Dirk Hartung, Lauritz Gerlach, Abhik Jana, and Michael J.~Bommarito II. 2023.
\newblock \href {https://arxiv.org/abs/2302.12039} {Natural language processing in the legal domain}.
\newblock \emph{Preprint}, arXiv:2302.12039.

\bibitem[{Kim et~al.(2021)Kim, Jiang, and Wang}]{kim_discovering_2021}
Seungbae Kim, Jyun-Yu Jiang, and Wei Wang. 2021.
\newblock \href {https://doi.org/10.1145/3437963.3441803} {Discovering undisclosed paid partnership on social media via aspect-attentive sponsored post learning}.
\newblock In \emph{Proceedings of the 14th {{ACM}} International Conference on Web Search and Data Mining}, {{WSDM}} '21, pages 319--327. Association for Computing Machinery.

\bibitem[{Kok-Shun and Chan(2025)}]{kok-shun_leveraging_2025}
Brice~Valentin Kok-Shun and Johnny Chan. 2025.
\newblock \href {https://doi.org/10.48550/arXiv.2502.15102} {Leveraging {{ChatGPT}} for sponsored ad detection and keyword extraction in {{YouTube}} videos}.
\newblock \emph{Preprint}, arXiv:2502.15102.

\bibitem[{Kolla et~al.(2024)Kolla, Salunkhe, Chandrasekharan, and Saha}]{kolla_llm-mod_2024}
Mahi Kolla, Siddharth Salunkhe, Eshwar Chandrasekharan, and Koustuv Saha. 2024.
\newblock \href {https://doi.org/10.1145/3613905.3650828} {{{LLM-mod}}: {{Can}} large language models assist content moderation?}
\newblock In \emph{Extended Abstracts of the {{CHI}} Conference on Human Factors in Computing Systems}, {{CHI EA}} '24, pages 1--8. Association for Computing Machinery.

\bibitem[{Li et~al.(2023)Li, Lin, Zhang, Fu, Chen, Lou, and Chen}]{li_making_2023}
Yifei Li, Zeqi Lin, Shizhuo Zhang, Qiang Fu, Bei Chen, Jian-Guang Lou, and Weizhu Chen. 2023.
\newblock \href {https://doi.org/10.48550/arXiv.2206.02336} {Making large language models better reasoners with step-aware verifier}.
\newblock \emph{Preprint}, arXiv:2206.02336.

\bibitem[{Lind(2014)}]{lind_significance_2014}
Douglas Lind. 2014.
\newblock \href {https://www.judges.org/news-and-info/the-significance-of-logic-for-law/} {The significance of logic for law. {{The National Judicial College}}}.

\bibitem[{Louis et~al.(2024)Louis, van Dijck, and Spanakis}]{louis_interpretable_2024}
Antoine Louis, Gijs van Dijck, and Gerasimos Spanakis. 2024.
\newblock \href {https://doi.org/10.1609/aaai.v38i20.30232} {Interpretable long-form legal question answering with retrieval-augmented large language models: 38th {{AAAI}} conference on artificial intelligence 2024}.
\newblock \emph{Proceedings of the 38th AAAI Conference on Artificial Intelligence}, 38:22266--22275.

\bibitem[{Martins et~al.(2022)Martins, Salles, Benevenuto, and Goussevskaia}]{martins_characterizing_2022}
Emanuelle~Azevedo Martins, Isadora Salles, Fabricio Benevenuto, and Olga Goussevskaia. 2022.
\newblock \href {https://doi.org/10.1145/3511095.3531289} {Characterizing sponsored content in facebook and instagram}.
\newblock In \emph{Proceedings of the 33rd {{ACM}} Conference on Hypertext and Social Media}, {{HT}} '22, pages 52--63. Association for Computing Machinery.

\bibitem[{Mathur et~al.(2018)Mathur, Narayanan, and Chetty}]{mathur_endorsements_2018}
Arunesh Mathur, Arvind Narayanan, and Marshini Chetty. 2018.
\newblock \href {https://doi.org/10.1145/3274388} {Endorsements on social media: {{An}} empirical study of affiliate marketing disclosures on {{YouTube}} and pinterest}.
\newblock \emph{Proc. ACM Hum.-Comput. Interact.}, 2.

\bibitem[{Medvedeva and Mcbride(2023)}]{medvedeva_legal_2023}
Masha Medvedeva and Pauline Mcbride. 2023.
\newblock \href {https://doi.org/10.18653/v1/2023.nllp-1.9} {Legal judgment prediction: {{If}} you are going to do it, do it right}.
\newblock In \emph{Proceedings of the Natural Legal Language Processing Workshop 2023}, pages 73--84. Association for Computational Linguistics.

\bibitem[{Mishra et~al.(2025)Mishra, Pathiraja, Parmar, Chidananda, Srinivasa, Liu, Payani, and Baral}]{mishra_investigating_2025}
Venkatesh Mishra, Bimsara Pathiraja, Mihir Parmar, Sat Chidananda, Jayanth Srinivasa, Gaowen Liu, Ali Payani, and Chitta Baral. 2025.
\newblock \href {https://doi.org/10.48550/arXiv.2502.05675} {Investigating the shortcomings of {{LLMs}} in step-by-step legal reasoning}.
\newblock \emph{Preprint}, arXiv:2502.05675.

\bibitem[{Palla et~al.(2025)Palla, García, Hauff, Fabbri, Damianou, Lindström, Taber, and Lalmas}]{palla_policy-as-prompt_2025}
Konstantina Palla, José Luis~Redondo García, Claudia Hauff, Francesco Fabbri, Andreas Damianou, Henrik Lindström, Dan Taber, and Mounia Lalmas. 2025.
\newblock \href {https://doi.org/10.1145/3715275.3732054} {Policy-as-prompt: {{Rethinking}} content moderation in the age of large language models}.
\newblock In \emph{Proceedings of the 2025 {{ACM}} Conference on Fairness, Accountability, and Transparency}, {{FAccT}} '25, pages 840--854. Association for Computing Machinery.

\bibitem[{Patel et~al.(2024)Patel, Kulkarni, Parmar, Budhiraja, Nakamura, Varshney, and Baral}]{patel_multi-logieval_2024}
Nisarg Patel, Mohith Kulkarni, Mihir Parmar, Aashna Budhiraja, Mutsumi Nakamura, Neeraj Varshney, and Chitta Baral. 2024.
\newblock \href {https://doi.org/10.48550/arXiv.2406.17169} {Multi-{{LogiEval}}: {{Towards}} evaluating multi-step logical reasoning ability of large language models}.
\newblock \emph{Preprint}, arXiv:2406.17169.

\bibitem[{Pałka et~al.(2025)Pałka, Lagioia, Liepina, Lippi, and Sartor}]{palka_make_2025}
Przemysław Pałka, Francesca Lagioia, Rūta Liepina, Marco Lippi, and Giovanni Sartor. 2025.
\newblock \href {https://doi.org/10.1007/s10506-025-09442-0} {Make privacy policies longer and appoint {{LLM}} readers}.
\newblock \emph{Artificial Intelligence and Law}.

\bibitem[{Plank(2022-12)}]{plank_problem_2022}
Barbara Plank. 2022-12.
\newblock \href {https://doi.org/10.18653/v1/2022.emnlp-main.731} {The “problem” of human label variation: {{On}} ground truth in data, modeling and evaluation}.
\newblock In \emph{Proceedings of the 2022 Conference on Empirical Methods in Natural Language Processing}, pages 10671--10682. Association for Computational Linguistics.

\bibitem[{Practice(2025)}]{practice_influencer_2025}
Advertising Standards Authority \{\textbackslash textbar\} Committee of~Advertising Practice. 2025.
\newblock \href {https://www.asa.org.uk/news/influencer-ad-disclosure-on-social-media-instagram-and-tiktok-report-2024.html} {Influencer ad disclosure on social media: {{Instagram}} and {{TikTok}} report (2024)}.

\bibitem[{Prasad et~al.(2023)Prasad, Saha, Zhou, and Bansal}]{prasad_receval_2023}
Archiki Prasad, Swarnadeep Saha, Xiang Zhou, and Mohit Bansal. 2023.
\newblock \href {https://doi.org/10.18653/v1/2023.emnlp-main.622} {{{ReCEval}}: {{Evaluating}} reasoning chains via correctness and informativeness}.
\newblock In \emph{Proceedings of the 2023 Conference on Empirical Methods in Natural Language Processing}, pages 10066--10086. Association for Computational Linguistics.

\bibitem[{Rogers et~al.(2023)Rogers, Balasubramanian, Derczynski, Dodge, Koller, Luccioni, Sap, Schwartz, Smith, and Strubell}]{rogers_closed_2023}
Anna Rogers, Niranjan Balasubramanian, Leon Derczynski, Jesse Dodge, Alexander Koller, Sasha Luccioni, Maarten Sap, Roy Schwartz, Noah~A. Smith, and Emma Strubell. 2023.
\newblock \href {https://hackingsemantics.xyz/2023/closed-baselines/} {Closed {{AI}} models make bad baselines}.

\bibitem[{Santos~Rodrigues et~al.(2021)Santos~Rodrigues, Munaro, and Paraiso}]{santos_rodrigues_identifying_2021}
João~P. Santos~Rodrigues, Ana~C. Munaro, and Emerson~Cabrera Paraiso. 2021.
\newblock \href {https://doi.org/10.1109/SMC52423.2021.9659291} {Identifying sponsored content in {{YouTube}} using information extraction}.
\newblock In \emph{2021 {{IEEE}} International Conference on Systems, Man, and Cybernetics ({{SMC}})}, pages 3075--3080.

\bibitem[{Shailya et~al.(2025)Shailya, Rajpal, Krishnan, and Ravindran}]{shailya_lext_2025}
Krithi Shailya, Shreya Rajpal, Gokul~S. Krishnan, and Balaraman Ravindran. 2025.
\newblock \href {https://doi.org/10.48550/arXiv.2504.06227} {{{LExT}}: {{Towards}} evaluating trustworthiness of natural language explanations}.
\newblock \emph{Preprint}, arXiv:2504.06227.

\bibitem[{Swart et~al.(2020)Swart, Lopez, Mathur, and Chetty}]{swart_is_2020}
Michael Swart, Ylana Lopez, Arunesh Mathur, and Marshini Chetty. 2020.
\newblock \href {https://doi.org/10.1145/3313831.3376178} {Is this an ad?: {{Automatically}} disclosing online endorsements on {{YouTube}} with {{AdIntuition}}}.
\newblock In \emph{Proceedings of the 2020 {{CHI}} Conference on Human Factors in Computing Systems}, {{CHI}} '20, pages 1--12. Association for Computing Machinery.

\bibitem[{Tan et~al.(2025)Tan, Zhuang, Montgomery, Tang, Cuadron, Wang, Popa, and Stoica}]{tan_judgebench_2025}
Sijun Tan, Siyuan Zhuang, Kyle Montgomery, William~Y. Tang, Alejandro Cuadron, Chenguang Wang, Raluca~Ada Popa, and Ion Stoica. 2025.
\newblock \href {https://doi.org/10.48550/arXiv.2410.12784} {{{JudgeBench}}: A benchmark for evaluating {{LLM-based}} judges}.
\newblock \emph{Preprint}, arXiv:2410.12784.

\bibitem[{Trautmann et~al.(2024)Trautmann, Ostapuk, Grail, Pol, Bonifazi, Gao, and Gajek}]{trautmann_measuring_2024}
Dietrich Trautmann, Natalia Ostapuk, Quentin Grail, Adrian Pol, Guglielmo Bonifazi, Shang Gao, and Martin Gajek. 2024.
\newblock \href {https://doi.org/10.18653/v1/2024.nllp-1.14} {Measuring the groundedness of legal question-answering systems}.
\newblock In \emph{Proceedings of the Natural Legal Language Processing Workshop 2024}, pages 176--186. Association for Computational Linguistics.

\bibitem[{Tyen et~al.(2024)Tyen, Mansoor, Cărbune, Chen, and Mak}]{tyen_llms_2024}
Gladys Tyen, Hassan Mansoor, Victor Cărbune, Peter Chen, and Tony Mak. 2024.
\newblock \href {https://doi.org/10.48550/arXiv.2311.08516} {{{LLMs}} cannot find reasoning errors, but can correct them given the error location}.
\newblock \emph{Preprint}, arXiv:2311.08516.

\bibitem[{Vats et~al.(2023)Vats, Zope, De, Sharma, Bhattacharya, Nigam, Guha, Rudra, and Ghosh}]{vats_llms_2023}
Shaurya Vats, Atharva Zope, Somsubhra De, Anurag Sharma, Upal Bhattacharya, Shubham~Kumar Nigam, Shouvik Guha, Koustav Rudra, and Kripabandhu Ghosh. 2023.
\newblock \href {https://doi.org/10.18653/v1/2023.findings-emnlp.831} {{{LLMs}} – the good, the bad or the indispensable?: A use case on legal statute prediction and legal judgment prediction on indian court cases}.
\newblock In \emph{Findings of the Association for Computational Linguistics: {{EMNLP}} 2023}, pages 12451--12474. Association for Computational Linguistics.

\bibitem[{Wang et~al.(2024)Wang, Li, Chen, Cai, Zhu, Lin, Cao, Kong, Liu, Liu, and Sui}]{wang_large_2024}
Peiyi Wang, Lei Li, Liang Chen, Zefan Cai, Dawei Zhu, Binghuai Lin, Yunbo Cao, Lingpeng Kong, Qi~Liu, Tianyu Liu, and Zhifang Sui. 2024.
\newblock \href {https://doi.org/10.18653/v1/2024.acl-long.511} {Large language models are not fair evaluators}.
\newblock In \emph{Proceedings of the 62nd Annual Meeting of the Association for Computational Linguistics (Volume 1: {{Long}} Papers)}, pages 9440--9450. Association for Computational Linguistics.

\bibitem[{Xu et~al.(2023-12)Xu, T.y.s.s, Ichim, Risini, Plank, and Grabmair}]{xu_dissonance_2023}
Shanshan Xu, Santosh T.y.s.s, Oana Ichim, Isabella Risini, Barbara Plank, and Matthias Grabmair. 2023-12.
\newblock \href {https://doi.org/10.18653/v1/2023.emnlp-main.594} {From dissonance to insights: {{Dissecting}} disagreements in rationale construction for case outcome classification}.
\newblock In \emph{Proceedings of the 2023 Conference on Empirical Methods in Natural Language Processing}, pages 9558--9576. Association for Computational Linguistics.

\bibitem[{Yin et~al.(2025)Yin, Laban, Peng, Zhou, Mao, Vats, Ross, Agarwal, Xiong, and Wu}]{yin_bingoguard_2025}
Fan Yin, Philippe Laban, Xiangyu Peng, Yilun Zhou, Yixin Mao, Vaibhav Vats, Linnea Ross, Divyansh Agarwal, Caiming Xiong, and Chien-Sheng Wu. 2025.
\newblock \href {https://doi.org/10.48550/arXiv.2503.06550} {{{BingoGuard}}: {{LLM}} content moderation tools with risk levels}.
\newblock \emph{Preprint}, arXiv:2503.06550.

\bibitem[{Yuan et~al.(2024)Yuan, Kao, Wu, Cheung, Chan, Cheung, Chan, and Chen}]{yuan_bringing_2024}
Mingruo Yuan, Ben Kao, Tien-Hsuan Wu, Michael M.~K. Cheung, Henry W.~H. Chan, Anne S.~Y. Cheung, Felix W.~H. Chan, and Yongxi Chen. 2024.
\newblock \href {https://doi.org/10.1007/s10506-023-09367-6} {Bringing legal knowledge to the public by constructing a legal question bank using large-scale pre-trained language model}.
\newblock \emph{Artificial Intelligence and Law}, 32(3):769--805.

\bibitem[{Zarei et~al.(2020)Zarei, Ibosiola, Farahbakhsh, Gilani, Garimella, Crespi, and Tyson}]{zarei_characterising_2020}
Koosha Zarei, Damilola Ibosiola, Reza Farahbakhsh, Zafar Gilani, Kiran Garimella, Noël Crespi, and Gareth Tyson. 2020.
\newblock \href {https://doi.org/10.1109/ASONAM49781.2020.9381309} {Characterising and detecting sponsored influencer posts on instagram}.
\newblock In \emph{2020 {{IEEE}}/{{ACM}} International Conference on Advances in Social Networks Analysis and Mining ({{ASONAM}})}, pages 327--331.

\bibitem[{Zhang et~al.(2024)Zhang, Li, Wu, Ai, Liu, Zhang, and Ma}]{zhang_evaluation_2024}
Ruizhe Zhang, Haitao Li, Yueyue Wu, Qingyao Ai, Yiqun Liu, Min Zhang, and Shaoping Ma. 2024.
\newblock \href {https://arxiv.org/abs/2403.11152v1} {Evaluation ethics of {{LLMs}} in legal domain. {{arXiv}}.org}.

\end{thebibliography}

\appendix

\section{Prompt template}

\label{appendix: base prompt}
\# Identity

You are a legal expert, as well as a social media content moderator who is responsible for keeping monetised posts compliant with the advertisement disclosure rules.
 
\# Context

You are reviewing social media posts that are likely to be undisclosed ads. Your goal is to determine, under Dutch advertising law, whether the post is in fact an advertisement -- regardless of whether disclosure is present. Disclosed posts should still be classified as ads if they meet the criteria. The classification is based on the nature of the post, not solely the presence/absence of disclosure.
 
\# Task

You're given these social media posts. Based on your legal knowledge of Dutch advertising law, decide if this post is an advertisement. First, justify your decision step-by-step using legal and contextual reasoning, referring to the specific articles from the regulations, and making a legal argument.
 
\# Output format

Please provide the following outputs, in this order, strictly adhering to the instructions and avoiding verbosity:

<Justification> Output the detailed reasoning that directed your result. This must be the chain-of-thought style legal reasoning, grounded in the Dutch Advertising Code and the Advertising Code Social Media \& Influencer Marketing.

<Is the post an advertisement> True (1)/False (0). Output strictly as 1 or 0.
 
Always decide the label only after completing the reasoning.


\section{Extra tables}
\label{appendix:unified}


\begin{landscape}
\begin{table}[ht]
\centering
\resizebox{\linewidth}{!}{%
\begin{tabular}{p{4cm} p{7cm} p{10cm}}
\hline
\textbf{Error Type} & \textbf{Description} & \textbf{Example} \\
\hline
Wrong interpretation of legal citations & The LLM gives an argument based on certain articles, but that is not what the article means & There is no disclosure or clear indication that this is a promotional post or part of an advertising campaign (Articles 11 and 3 of Dutch Advertising Code). Article 3: Advertising may not be contrary to the general interest, public order, or morality. \\
No citation & Explanations don't include any legal citations & --- \\
Citation is not clear & It cites multiple articles but didn't clearly map them out & There is no disclosure or clear indication that this is a promotional post or part of an advertising campaign (Articles 11 and 3 of the Dutch Advertising Code and RSM). It didn't name specifically which article is from which code. \\
Hallucinations on the legal citations & When the answer includes legal information that is not in the regulation & According to Article 7, but there is actually no Article 7. According to (Some random law that you can check on Google if it really exists). \\
Hallucinations on the content & Besides legal content, the answer includes content that doesn't exist, such as the brand name & The influencer cooperate with @Nike, but actually there is no mentioning of Nike at all in the original post. \\
Mistaken potential cues & Don't/Wrongly identify a clue as advertisements or advertisers & \#fyp is not an ad cue, but LLM believe it is; @a friend, but recognises that as an advertiser. Find the potential clues (\#Nike), but don't take them as the evidence. \\
Reasoning ends up opposite the output & The reasoning process is opposite to the final conclusion. It means trying to reason it as an ad, but the final conclusion said it is not & Is there a Relevant Relationship? – Yes, the post explicitly mentions collaboration with @thewoolmarkcompany, indicating a business relationship. This relationship influences the content, as the post promotes wool products, possibly as part of sponsored content. With explanations all like this, it still label the post as False (non-ad). \\
\hline
\end{tabular}%
}
\caption{Types of errors in LLM responses regarding advertising identification.}
\label{tab:llm_errors}
\end{table}
\end{landscape}

\label{appendix: matrix tables}

\begin{table*}[ht]
\centering
\begin{tabular}{lllrrr}
\toprule
 & Model & Prompting strategy & Precision & Recall & F1 Score \\
\midrule
0 & logistic regression (TF-IDF)& & 0.60  &  0.55   &   0.57 \\
1 & gemini-2.5-flash-lite & no\_article          & 0.72 & 0.84 & 0.77 \\
2 & gemini-2.5-flash-lite & article              & 0.74 & \textbf{0.88} & \textbf{0.80} \\
3 & gemini-2.5-flash-lite & article\_explanation & 0.75 & 0.86 & \textbf{0.80} \\
4 & gpt-5-nano            & no\_article          & \textbf{0.80} & 0.79 & \textbf{0.80} \\
5 & gpt-5-nano            & article              & 0.75 & 0.70 & 0.73 \\
6 & gpt-5-nano            & article\_explanation & 0.76 & 0.61 & 0.68 \\
\bottomrule
\end{tabular}
\caption{Comparison of performance across models and prompting strategies for the ambiguous cases in the task of advertisement identification}
\label{tab:performance for tricky cases}
\end{table*}


\begin{table*}[ht]
\centering
\resizebox{\textwidth}{!}{%
\begin{tabular}{llllrlrlr}
\toprule
 & Model & Prompting strategy & Category & Accuracy & Category & Accuracy & Category & Accuracy \\
\midrule
0 & logistic regression (TF-IDF) &          & Disclosed ads & 0.92 & Hidden ads & 0.86 & Organic & 0.73 \\
1 & gemini-2.5-flash-lite & no\_article          & Disclosed ads & \textbf{0.94} & Hidden ads & 0.91 & Organic & 0.86 \\
2 & gemini-2.5-flash-lite & article             & Disclosed ads & \textbf{0.94} & Hidden ads & \textbf{0.93} & Organic & 0.88 \\
3 & gemini-2.5-flash-lite & article\_explanation & Disclosed ads & 0.93 & Hidden ads & 0.91 & Organic & 0.88 \\
4 & gpt-5-nano            & no\_article          & Disclosed ads & 0.92 & Hidden ads & 0.87 & Organic & 0.90 \\
5 & gpt-5-nano            & article             & Disclosed ads & 0.90 & Hidden ads & 0.79 & Organic & \textbf{0.92} \\
6 & gpt-5-nano            & article\_explanation & Disclosed ads & 0.89 & Hidden ads & 0.76 & Organic & \textbf{0.92} \\
\bottomrule
\end{tabular}%
}
\caption{Accuracy by model, prompting strategy, and type of content.}
\label{tab:content_category_accuracy}
\end{table*}

\begin{table*}[ht]
\centering
\resizebox{\textwidth}{!}{%
\begin{tabular}{llllrrrrrrr}
\toprule
 & Model & Variant & Data Source & e1 (\%) & e2 (\%) & e3 (\%) & e4 (\%) & e5 (\%) & e6 (\%) & e7 (\%) \\
\midrule
0 & gemini-2.5-flash-lite & article & disclosed\_ads & 0.00 & 4.35 & 4.35 & 0.00 & 0.00 & 0.00 & 0.00 \\
1 & gemini-2.5-flash-lite & article & organic & 3.85 & 7.69 & 11.54 & 0.00 & 7.69 & 11.54 & 0.00 \\
2 & gemini-2.5-flash-lite & article & undisclosed\_ads & 14.29 & 4.76 & 4.76 & 0.00 & 14.29 & 28.57 & 4.76 \\
3 & gemini-2.5-flash-lite & article\_explanation & disclosed\_ads & 0.00 & 4.35 & 0.00 & 0.00 & 0.00 & 0.00 & 0.00 \\
4 & gemini-2.5-flash-lite & article\_explanation & organic & 11.54 & 0.00 & 7.69 & 0.00 & 3.85 & 19.23 & 0.00 \\
5 & gemini-2.5-flash-lite & article\_explanation & undisclosed\_ads & 9.52 & 4.76 & 4.76 & 0.00 & 0.00 & 9.52 & 0.00 \\
6 & gemini-2.5-flash-lite & no\_article & disclosed\_ads & 17.39 & 78.26 & 26.09 & 8.70 & 0.00 & 0.00 & 0.00 \\
7 & gemini-2.5-flash-lite & no\_article & organic & 3.85 & 92.31 & 23.08 & 0.00 & 3.85 & 15.38 & 0.00 \\
8 & gemini-2.5-flash-lite & no\_article & undisclosed\_ads & 33.33 & 71.43 & 28.57 & 9.52 & 4.76 & 14.29 & 0.00 \\
9 & gpt-5-nano & article & disclosed\_ads & 13.04 & 4.35 & 39.13 & 0.00 & 0.00 & 0.00 & 0.00 \\
10 & gpt-5-nano & article & organic & 7.69 & 0.00 & 3.85 & 0.00 & 0.00 & 7.69 & 0.00 \\
11 & gpt-5-nano & article & undisclosed\_ads & 14.29 & 0.00 & 19.05 & 4.76 & 4.76 & 4.76 & 0.00 \\
12 & gpt-5-nano & article\_explanation & disclosed\_ads & 4.35 & 0.00 & 39.13 & 0.00 & 0.00 & 0.00 & 0.00 \\
13 & gpt-5-nano & article\_explanation & organic & 7.69 & 0.00 & 26.92 & 3.85 & 0.00 & 7.69 & 0.00 \\
14 & gpt-5-nano & article\_explanation & undisclosed\_ads & 9.52 & 4.76 & 28.57 & 0.00 & 0.00 & 4.76 & 0.00 \\
15 & gpt-5-nano & no\_article & disclosed\_ads & 0.00 & 95.65 & 26.09 & 4.35 & 0.00 & 0.00 & 0.00 \\
16 & gpt-5-nano & no\_article & organic & 3.85 & 76.92 & 42.31 & 7.69 & 0.00 & 7.69 & 0.00 \\
17 & gpt-5-nano & no\_article & undisclosed\_ads & 4.76 & 61.90 & 38.10 & 9.52 & 4.76 & 0.00 & 0.00 \\
\bottomrule
\end{tabular}%
}
\caption{Error percentages by model, prompting strategy, and data source. Each value represents the proportion of posts containing the corresponding error. (e1) \textit{Wrong interpretation of legal citations}, (e2) \textit{No citation}, (e3) \textit{Citation is not clear}, (e4) \textit{Hallucinations on the legal citations}, (e5) \textit{Hallucinations on the content}, (e6) \textit{Mistaken potential cues}, and (e7) \textit{Reasoning results in opposite output}. Each value represents the proportion of posts exhibiting the corresponding error. }
\label{tab:errors by content}
\end{table*}

\end{document}